\documentclass{article}
\usepackage{mlspconf,amsmath,graphicx,mystyle,epstopdf,url}
\usepackage[ruled]{algorithm2e}
\usepackage{mathtools}

\def\mmx{{\alpha}}
\def\MM{{\beta}}
\def\xgt{{x_0}}

\def\n{{\tilde n}}
\def\d{{\mathbf d}}

\newcommand{\vertiii}[1]{{\left\vert\kern-0.25ex\left\vert\kern-0.25ex\left\vert #1 
		\right\vert\kern-0.25ex\right\vert\kern-0.25ex\right\vert}}

\title{COVER TREE COMPRESSED SENSING FOR FAST MR FINGERPRINT RECOVERY}
\name{Mohammad Golbabaee*, Zhouye Chen$^\dagger$, Yves Wiaux$^\dagger$ and Mike E. Davies*
	\thanks{This work is partly funded by the EPSRC grant EP/M019802/1 and the ERC C-SENSE project (ERC-ADG-2015-694888). 
		}
	}
\address{*Institute for digital communications, University of Edinburgh, EH9 3JL, UK\\
$^\dagger$Institute of sensors, signals and systems, Heriot-Watt University, EH14 4AS, UK}
\begin{document}
\ninept
\maketitle
\begin{abstract}
We adopt a data structure in the form of cover trees and iteratively apply approximate nearest neighbour (ANN) searches for fast compressed sensing reconstruction of signals living on discrete smooth manifolds. 
Leveraging on the recent stability results for the inexact Iterative Projected Gradient (IPG) algorithm and by using the cover tree's ANN searches, 
we decrease the projection cost of the IPG algorithm to be 
logarithmically growing with data population for low dimensional smooth manifolds. We apply our results to quantitative MRI compressed sensing and in particular within the Magnetic Resonance Fingerprinting (MRF) framework. For a similar (or sometimes better) reconstruction accuracy, we report 2-3 orders of magnitude reduction in computations compared to the standard iterative method, which uses brute-force searches.
\end{abstract}
\section{Introduction}

Compressed sensing (CS) algorithms adopt efficient signal models to achieve accurate reconstruction given small number of non-adaptive measurements~\cite{DonohoCS, CRT:CS}. CS consists of a linear sampling model:
\eql{
	\label{eq:CSsampling}
	y \approx Ax_0,
}
where  $y$ is a noisy $m$-dimensional vector of measurements taken from the ground truth signal $x_0$ in a high dimension $n\gg m$ through using a linear mapping $A$. The success of CS for stably inverting such ill-posed systems relies on the fact that natural signals often follow low-dimensional models $\Cc$ with limited degrees of freedom compared to their original ambient dimension, and that certain random sampling schemes can compactly (in terms of the measurements) preserve the key signal information (see e.g.~\cite{RichCSreview} for an overview on different CS models and their corresponding sample complexities). 
Variational approach for CS reconstruction consists of solving e.g. the following constrained least-square problem:
\eql{\label{eq:opt1}
	\hat x = \argmin_{x \in \Cc} \norm{y-Ax}_2^2.
}  
where the constraint takes into account the signal model.


First order methods based on iterative local gradient and projection (or also proximal) updates 
are popular for CS recovery due to their scalability for big data  problems~\cite{Volkan:bigdata} and their flexibility in handling various complicated (convex/nonconvex and sometimes combinatorial) models~\cite{Blumen}. Starting from an initial point e.g. $x^{k=0}=\mathbf{0}$  and for a given step size $\mu$, the solution of Iterative Projected Gradient (IPG)  iteratively updates as follows:
\eql{\label{eq:IPGD}
	x^{k+1} = \Pp_{\Cc} \left( x^{k} - \mu A^H(Ax^{k}-y) \right), 
}
where the projection is defined as $\Pp_\Cc(x)\in \argmin_{u\in\Cc}\norm{x-u}$ and throughout, $\norm{.}$ stands for the Euclidean norm.

%
%

In this work we consider compressed sensing of signals living on a low-dimensional smooth manifold. In this case, the model projection might not be computationally easy for certain complex manifolds. A popular approach consists of collecting discrete samples of the manifold in a dictionary~\cite{recht:discretize,SCOOP}, i.e. a \emph{data driven} CS approach,  and approximate the projection step by searching the nearest atom in this dictionary.  Depending on the manifold complexity and the desired approximation accuracy, the size of this dictionary can be very large in practice. In this case performing an \emph{exhaustive} nearest neighbour (NN) search could become computationally challenging. 

We address this shortcoming by preprocessing the dictionary and build a tree structure suitable for fast \emph{approximate} nearest neighbour (ANN) searches. Recently, it as been shown 
that provided enough measurements and for moderate approximations the \emph{inexact} IPG achieves similar solution accuracy as for the exact algorithm~\cite{me:inexactIPG}. 
In this regard, we use the \emph{cover tree} structure~\cite{beygelzimer2006cover,Navigating} which is proposed for fast ANN searches on large datasets with low intrinsic dimensionality and with a search complexity $O(\log(d))$ in time, where $d$ is the dataset population (in contrast with the linear complexity of an exact exhaustive search $O(d)$). 

We apply our results to accelerate the Magnetic Resonance Fingerprint (MRF) recovery from under-sampled k-space measurements; an emerging field of study in computational medical imaging~\cite{MRF}. 
%
Since the MR fingerprints belong to the low-dimensional manifold of Bloch dynamic equations, we show that using the cover tree ANN search significantly shortcuts the computations compared to an exact IPG with exhaustive searches as in~\cite{BLIPsiam}.  

%

\section{Preliminaries}
The following embedding assumption plays a critical role in stability analysis of the (convex/nonconvex) IPG algorithm~\cite{BW:manifold,Blumen,me:inexactIPG}:
{\defn{\label{def:bilip} $A$ is bi-Lipschitz w.r.t. $\Cc$, if $\forall x,x'\in \Cc$ there exists constants $0<\alpha\leq \beta$ such that
		\eql{\label{eq:bilip}
			\alpha \norm{x-x'}^2 \leq \norm{A(x-x')}^2\leq \beta \norm{x-x'}^2  }	
	}}
Dealing with complicated signal models, the projection step can bring a huge computational burden at each iteration, and thus a natural line of thought is to perform this step with cheaper (available) approximations. The inexact IPG takes the following form:
\eql{\label{eq:inIP2}
	x^k = \Pp_\Cc^{\epsilon}\left(x^{k-1}-\mu A^H(Ax^{k}-y) \right).}
We consider the class of $(1+\epsilon)$-approximate projections: for any $\epsilon\geq0$:
\eql{\label{eq:eproj}
	\Pp_\Cc^{\epsilon}(x) \in \Big\{ u\in \Cc :\,	\norm{u-x} \leq (1+\epsilon)\norm{\Pp_\Cc(x)-x}  \Big\}. 
}
The following result states that if the embedding holds (with a better conditioning compared to the exact IPG case $\epsilon=0$) then the inexact algorithm  still achieves a solution accuracy comparable to the exact IPG~\cite{me:inexactIPG}:
{\thm{\label{th:inexactLS2} Assume $(A,\Cc)$ is bi-Lipschitz, $\xgt\in \Cc$ and that
		\eq{\sqrt{\epsilon+\epsilon^2}\leq \delta\sqrt{\mmx}/{\vertiii{A}} \qandq \MM < (2-2\delta+\delta^2) \mmx} 	
		for $\epsilon\geq 0$ and some constant $\delta \in [0,1)$.
		Set the step size $\left((2-2\delta+\delta^2) \mmx\right)^{-1}<\mu\leq\MM^{-1}$. The sequence generated by Algorithm \eqref{eq:inIP2} obeys the following bound:
		\eq{
			\norm{x^{k}-\xgt}\leq  \rho^k \norm{\xgt} + 
			\frac{\kappa_w}{1-\rho}w
		}
		where 
		\begin{align*}
		\rho=\sqrt{\frac{1}{\mu \mmx} -1}+ \delta, \quad
		\kappa_w= 2\frac{\sqrt{\MM}}{\mmx}+\sqrt\mu\delta, \quad
		w=\norm{y-A \xgt}.
		\end{align*}		
		
	}} 	
	{\rem{Increasing $\epsilon$ slows down the rate $\rho$ of linear convergence, however the iterations are potentially cheaper. Too large approximations $\epsilon$ may result in divergence.}}
	{\rem{After a finite $K=O(\log(\tau^{-1}))$ number of iterations Algorithm \eqref{eq:inIP2} achieves the solution accuracy $\norm{x^K-x_0}=O(w)+\tau$, for any  $\tau>0$.}}

\section{Data driven CS in product space}
\label{sec:datadrivenCS}

Data driven CS corresponds to cases where in the absence of an algebraic (semi-algebraic) physical model one resorts to collecting a large number of data samples in a dictionary and use it e.g. as a \emph{point cloud} model for CS reconstruction~\cite{RichCSreview}.  
Data driven CS finds numerous applications e.g. in hyperspectral imaging (HSI)~\cite{TIPHSI}, mass spectroscopy (e.g. MALDI imaging) \cite{Kobarg2014} and MRF recovery~\cite{MRF,BLIPsiam} just to name a few. For instance the USGS Hyperspectral library\footnote{\url{http://speclab.cr.usgs.gov}} contains the spectral signatures of thousands of substances. This side-information is shown to enhance CS recovery and classification~\cite{TIPHSI,meEUSIPCO10}. Here the model is discrete and nonconvex:
\eql{\label{eq:dmodel}
	\widetilde \Cc:=\bigcup_{i=1}^{d}\{\psi_i\} \in \CC^{\n}}
and it corresponds to atoms of a (possibly complex-valued) dictionary $\Psi$ customized for e.g. HSI, MALDI or MRF data.




A multichannel image such as HSI, MALDI or MRF can be represented by a $\n\times J$ matrix $X$, where $n=\n J$ is the total number of pixels, $J$ is the spatial resolution and $\n$ is the number of channels. 
Under the \emph{pixel purity} assumption each spatial pixel corresponds to a certain material with specific signature. Considering a notion of \emph{signal intensity} in this model (e.g. due to variable illumination conditions in HSI, or a non-uniform proton density in MRF), the columns of $X$ belong to the \emph{cone} associated with the signatures~\eqref{eq:dmodel}. Denoting by $X_j$ the $j$th column of $X$ (i.e. a multi-dimensional spatial pixel), we have 
\eql{\label{eq:immodel}
	X_{j}\in  \text{cone}(\widetilde \Cc), \quad \forall j=1,\ldots,J,
} 
where the cone is also a discrete set defined as
\eq{
	\text{cone}(\widetilde \Cc):=\{x\in \CC^{\n}: x/\gamma \in \widetilde \Cc \,  \text{for some} \, \gamma>0 \}.
}	
Here, $\gamma$ corresponds to the signal intensity (per spatial pixel).

The CS sampling and reconstruction follows \eqref{eq:CSsampling} and \eqref{eq:opt1} by setting $x^*:=X_\text{vec}$ (by $X_\text{vec}$ we denote the vector-rearranged form of the matrix $X$) and the only update is that now  
the solution lives in a \emph{product space} of the same model i.e. 
\eql{
	\label{eq:prod}\Cc:=\prod_{j=1}^J \text{cone}(\widetilde\Cc)
} 

In a single dimensional setting i.e. $\Cc=\text{cone}(\widetilde \Cc)$,  
  there will be no need for an IPG algorithm to solve problem \eqref{eq:opt1}. If the embedding holds with a $\mmx>0$ (i.e. an identifiability condition) then the atom with a maximum coherence $\langle A \psi_i/\norm{A \psi_i},y\rangle$ with the measurements identifies the correct signature in a \emph{single} iteration. 
However such a direct approach in a multi-dimensional setting $J\gg1$ 
 (e.g. when the sampling model non-trivially combines columns of $X$) generally has an \emph{exponential} $O(d^J)$ complexity because of the combinatorial nature of the product space constraints. A tractable scheme which has been frequently considered for this case e.g. in \cite{BLIPsiam} would be an IPG type algorithm~\eqref{eq:IPGD}.
 
 After the gradient update $Z_{\text{vec}}$ = $X_{\text{vec}}^k + \mu A^H(y-AX_{\text{vec}}^k)$, the projection onto the product model \eqref{eq:prod} decouples into separate cone projections for each spatial pixel $j$. The cone projection $\Pp_{\text{cone}(\widetilde \Cc)}(Z_j)$ follows the steps below: 
 \begin{align}\label{eq:exactP}
 &i^* = \argmin_i \norm{Z_j -\psi_i/\norm{\psi_i}} &\text{(NN search)}\\
 &X^{k+1}_j = \Pp_{\text{cone}(\widetilde \Cc)}(Z_j) = \gamma_j \psi_{i^*}& \text{(rescaling)}
 \end{align}
 where,  $\gamma_j = \max\left( \text{real}(\langle Z_j ,\psi_{i^*}\rangle)/  {\norm{\psi_{i^*}}^2},0\right)$ is the per-pixel signal intensity.
 
 
 
%
%
As a result, the IPG algorithm breaks down the computations into local updates namely, the gradient and projection steps for which an exact projection step e.g. by using a brute-force nearest neighbour (NN) search, has complexity $O(Jd)$ in time.

\section{Cover tree for fast nearest neighbour search}

With the data driven CS formalism and discretization of the model  
the projection step of IPG reduces to a search for the nearest atom in each of the product spaces, however in a potentially very large $d$ size dictionary. 
Thus search strategies with linear complexity in $d$, e.g. an exhaustive search, can be a serious bottleneck for solving such problems. 
A very well-established approach to overcome the complexity of an exhaustive NN search on a large dataset 
consists of hierarchically partitioning the solution space and forming a \emph{tree} structure whose nodes representing those partitions, and then using a  branch-and-bound method on the resulting tree for a fast Approximate NN (ANN) search with $o(d)$ complexity. 
In this regard, we address the computational shortcoming of the projection step in the exact IPG by preprocessing $\widetilde \Cc$ and form a \emph{cover tree} structure suitable for fast ANN searches \cite{beygelzimer2006cover,Navigating}.

A cover tree is a levelled tree whose nodes (i.e. associated data points) at different scales form covering nets for data points at multiple resolutions. Denote by $\Ss_l$ the set of nodes appearing at scale $l=1,\ldots,L_{\max}$ and by $\sigma:=\max_{\psi\in\widetilde \Cc}\norm{\psi_\text{root}-\psi}$ the maximal coverage by the root, then a cover tree structure must have the following  three properties:
\begin{enumerate}
	\item Nesting: $\Ss_l \subseteq \Ss_{l+1}$, once a point $p$ appears as a node in $\Ss_l$, then every lower level in the tree has that node.
	\item Covering: every node $q\in \Ss_{l+1}$ has a parent node  $p\in \Ss_{l}$, where $\norm{p-q}\leq \sigma2^{-l} $. As a result, covering becomes finer at higher scales in a dyadic fashion. 
	
	\item Separation: nodes belonging to the same scale are separated by a minimal distance which dyadically shrinks at higher scales i.e. $\forall q,q'\in\Ss_l$ we have $\norm{q-q'}>\sigma2^{-l}$.   
\end{enumerate}  
Each node $p$ also keeps the maximum distance to its descendants denoted by
$\mathrm{maxdist}(q):= \max_{q'\in \text{descendant}(q)} \norm{q-q'}\leq \sigma2^{-l+1}$. 
which will be useful for fast NN/ANN search.\footnote{Note that any node $q\in \Ss_l$ due to the covering property satisfies 
$maxdist(q) \leq \sigma \left( 2^{-l}+2^{-l-1}	+ 2^{-l-2}+\dots \right)
< \sigma 2^{-l+1}$ and thus, one might avoid saving $maxdist$ values and use this upper bound instead.}
The \emph{explicit} tree representation requires the storage space $O(d)$~\cite{beygelzimer2006cover}.

\begin{algorithm}[t]
	\label{alg:findNN}
	\SetAlgoLined
	$Q_0 = \{\Ss_0\}$, $\Ss_0$ the root of $\Tt$, 
	$\d_{\min}=\norm{p-q_c}$\\
	$l=0$\\
	\While{$l<L_{\max}$ \, \text{AND} \, $\sigma 2^{-l+1}(1+\epsilon^{-1})> \d_{\min}$\,}
	{
		$Q=\left\{\text{children}(q):\, q\in Q_l \right\} $\\
		$q^* = \argmin_{q\in Q} \norm{p-q}$, \quad $\d = \norm{p-q^*}$ \\
		\If{$\d<\d_{\min}$}{$\d_{\min}=\d, \quad q_c=q^*$}
		$Q_{l+1} = \left\{q\in Q: \norm{p-q}\leq \d_{\min} + \mathrm{maxdist}(q) \right\}$\\
		$l  = l+1$\\
	}	
	\Return $q_c$\\
	\caption{\label{alg:NN} \textbf{$(1+\epsilon)$-ANN}(cover tree $\Tt$, query point $p$, current estimate $q_c \in \widetilde \Cc$),\quad  accuracy parameter $\epsilon$}
\end{algorithm}

{\defn{\label{def:eNN}Given a query $p$, 
		$q\in\widetilde \Cc$ is a $(1+\epsilon)$-ANN of $p$ for an $\epsilon\geq 0$ if it holds:
		$\norm{p-q} \leq (1+\epsilon)\inf_{u\in\widetilde \Cc}\norm{p-u}.$
			}}	
\newline
			
Algorithm \ref{alg:NN} details the branch-and-bound procedure for $(1+\epsilon)$-ANN search on a given cover tree (proof of correctness is available  in~\cite{beygelzimer2006cover}).
In short, we iteratively traverse down the cover tree and at each scale we populate the set of \emph{candidates} $Q_l$ with nodes which could be the ancestors of the solution and discard others (this refinement uses the triangular inequality and the lower bound on the distance of the grandchildren of $Q$ to $p$, based on  $\mathrm{maxdist}(q)$).  At the finest scale (before stopping) we search the whole set of final candidates and report an $(1+\epsilon)$-ANN point. Note that at each scale we only compute distances for non self-parent nodes (we pass, without computation, distance information of the self-parent children to finer scales). 
The case $\epsilon=0$ refers to the exact tree NN search where one has to continue  Algorithm~\ref{alg:NN} until the finest level of the tree.   One should distinguish between this strategy and performing a brute-force search. Although they both perform an exact NN search, the complexity of Algorithm~\ref{alg:NN} 
is empirically shown to be way less in practical datasets.

Although the cover tree construction is blind to the explicit structure of data, several key growth properties such as the tree's explicit depth, the number of children per node, and importantly the overall search complexity are characterized by the intrinsic \emph{doubling dimension} of the model~\cite{Navigating}. 
Practical datasets are often assumed to have small doubling dimensions e.g. when $\widetilde \Cc \subseteq \Mm$  
samples a $K$-dimensional manifold $\Mm$ with certain  regularity, one has $\dim_D(\widetilde \Cc)\leq \dim_D(\Mm)=O(K)$ \cite{dasgupta2008}. 
The following theorem bounds the complexity of  $(1+\epsilon)$-ANN cover tree search~\cite{Navigating,beygelzimer2006cover}: 
		{\thm{\label{thm:NNcomp2}Given a query which might not belong to $\widetilde \Cc$, the approximate $(1+\epsilon)$-ANN search on a cover tree takes at most 
				\eql{
					2^{O(\dim_D(\widetilde \Cc))}\log \Delta+(1/\epsilon)^{O(\dim_D(\widetilde \Cc))} 
				}
				computations in time with $O(\#\widetilde \Cc)$ memory requirement, where  $\Delta$ is the aspect ratio\footnote{The aspect ratio of a set $\Cc$ is $\Delta:=\frac{\max \norm{q-q'}}{\min \norm{q-q'}}$, $\forall q\neq q'\in \Cc$.} of $\widetilde \Cc$  and $\#$ stands for set cardinality.}	
		}
		
		For most applications $\log (\Delta) = O(\log(d))$~\cite{Navigating} and thus for datasets with low dimensional structures i.e. $\dim_D=O(1)$ and by using  moderate approximations one achieves a search complexity logarithmic in $d$, as opposed to the linear complexity of a brute-force search.
		
		Note that the complexity of an \emph{exact} cover tree search could be arbitrary high i.e. linear in $d$, at least in theory (unless the query belongs to the dataset~\cite{beygelzimer2006cover} which does not generally apply e.g. for the intermediate steps of the IPG). 
		However in our experiments we empirically observe  that the complexity of an exact cover tree NN is still much lower than performing an exhaustive search.

\subsection{Inexact data driven IPG algorithm}
We use cover tree's $(1+\epsilon)$-ANN approximate search to efficiently implement the cone projection step of the data driven IPG algorithm. 
We only replace the exact search step \eqref{eq:exactP} with the following approximation:
\eql{\label{eq:ANN1}
	\psi_{i^*} = \textbf{$(1+\epsilon)$-ANN}\left( \Tt,\frac{Z_j}{\norm{Z_j}}, \frac{X_j^k}{\gamma_j}\right), \quad \forall j=1,\ldots,J.
}
We denote by $\Tt$ the cover tree structure built for the normalized dictionary with atoms $\psi_i/\norm{\psi_i}$. For the current estimate we use the atom previously selected i.e. $X_j^k/\gamma_j$ for all $j$. We additionally normalize the gradient update $Z_j$ for the search input. This step may sound redundant however since all dictionary atoms are normalized and live on the  hypersphere, we empirically observed that this trick leads to a better  acceleration.

As a result by this update we obtain a $(1+\epsilon)$-approximate cone projection denoted by $\Pp^\epsilon_{\text{cone}(\widetilde \Cc)}(Z_j)$ satisfying definition \eqref{eq:eproj}. Therefore supported with an embedding assumption w.r.t. to the product model \eqref{eq:prod} and according to Theorem~\ref{th:inexactLS2} we can guarantee stable CS recovery using this algorithm with a small computational effort.

\section{Application in Quantitative MRI}


In \emph{quantitative MRI} 
rather than simply forming a single MR image with only  contrast information (i.e. qualitative MRI), physicists
are interested in measuring the NMR properties of tissues   namely, the $T1,T2$ relaxation times, $\delta f$ magnetic resonance and proton density (PD), and use these parameters to differentiate different biological tissues~\cite{toftsqmri}. The standard approach for parameter estimation is to acquire a large sequence
of images in different times from a simple excitation pulse (i.e. excitations in terms of rotating the magnetic field)  and then use an exponential model curve fitting to recover the decay exponents $T1,T2$ for each voxel. This procedure runs separately to estimate each relaxation time.
The long process of acquiring multiple fully sampled images, brings serious limitation to standard approaches in quantitative imaging to apply within a reasonable time and with an acceptable signal-to-noise ratio (SNR) and resolution.
%
%

\subsection{Magnetic Resonance Fingerprinting}

Recently, a novel parameter estimation process coined as the \emph{Magnetic Resonance Fingerprinting} (MRF) has been proposed to address this shortcoming. 
For the acquisition, MRF applies a shorter sequence of random excitations (i.e. $\Gamma \in [0,\pi/2]^{\n}$ random flip angels with  typically $\n\approx 300-1000$ and the time interval $TR\sim 10-40$ msec) for estimating all parameters $T1,T2,\delta f$, and proton density at once.
 After each excitation pulse the response is recorded  through the 
measurements taken from a small portion of k-space.
As a result the acquisition time can significantly decreases to a couple of minutes \cite{MRF}.   

The exponential model as for the traditional acquisition sequences will no longer hold for the MRF model. Due to using random excitations, the signatures casts a non-trivial implicit expression  readable by solving the \emph{Bloch dynamic equations} $\Bb(\Gamma,\Theta)\in \CC^\n $ customized for the set of parameters $\Theta=\{T1,T2,\delta f\}$. 
The MRF problem is an example of data driven CS model with an algebraic signal model (i.e. the Bloch equations) but a non-trivial projection. For this problem one constructs off-line a
dictionary of \emph{fingerprints} (i.e. magnetization responses) for all possible $d=\#\{\Theta_i\}_i$ parameters presented in normal tissues where the atoms $\psi_i:=\Bb(\Gamma,\Theta_i)$ uniquely represent the underlying set of parameters $\Theta_i$ . This corresponds
to sampling a low-dimensional manifold associated with the
solutions of the Bloch dynamic equations~\cite{BLIPsiam}, which for complicated excitation patterns neither the response nor the projection has an
analytic solution.


\subsection{MRF image model and parameter estimation}

An MRF image can be represented by a matrix $X\in \CC^{ \n\times J}$. Each column $X_j$ represents a $\n$-dimensional spatial pixel $j$ and each row $X^l$  represents a temporal image slice (with the spatial resolution equal to $J$) corresponding to the measurements taken after the excitation pulse $\Gamma_l$, $\forall l=1,\ldots,\n$. The total spatio-temporal resolution of the MRF image is $n:=\n J$. By the pixel purity assumption, the image follows the product space data driven model \eqref{eq:dmodel} and \eqref{eq:immodel} where each column/pixel $j$ correspond to a single fingerprint $\psi$ with a unique parameter  $\Theta$. The cone($\widetilde \Cc$) corresponds to the fingerprints dictionary and $\rho>0$ represents the proton density.\footnote{The proton density can be generally a complex number for which only the rescaling step of the cone projection slightly changes and the rest of the analysis similarly holds, for more details see~\cite{BLIPsiam}. 
	} 
The MRF acquisition model is linear as follows:
\eql{\label{eq:CSMRF}	
	Y^l \approx \Ff_{\Omega_l} (X^l), \quad \forall l=1,\ldots \n,
}
where  $\Ff$ denotes the two-dimensional Fourier transform, and $\Ff_{\Omega_l}: \CC^J \rightarrow \CC^{m/\n}$ subsamples the k-space coefficients of the  image slice $X^l$ according a pattern/set $\Omega_l$.  The k-space subsampling could correspond to a spiral pattern~\cite{MRF} or a random horizontal/vertical line subselection~\cite{BLIPsiam} e.g. in the Echo Planar Imaging (EPI). 
By vectorization of $X,Y$, we have a product space CS model similar to~\eqref{eq:CSsampling} for a sampling protocol $A$ acting spatially however with different k-space sampling pattern per image slice. 


The CS reconstruction proposed by~\cite{BLIPsiam} is based on a data driven IPG algorithm (with the \emph{exact} cone projection described in Section~\ref{sec:datadrivenCS}) where one by identifying the correct fingerprint per pixel $X_j$ can recover the underlying parameters $\Theta$ stored in a look-up table (the last rescaling step also recovers the proton density $\gamma_j$). 
It has been shown in~\cite[Theorem 1]{BLIPsiam} that if $\Omega_l$ independently at random subselects the rows (or columns) of the k-space, which corresponds to a \emph{random EPI sampling protocol}, the resulting forward map $A$ is bi-Lipschitz  w.r.t. the product model $\Cc=\prod_{j=1}^J \text{cone}(\widetilde \Cc)$, provided sufficient \emph{discriminations} between the fingerprints.\footnote{One achieves a suitable discrimination between the MR fingerprints by e.g. choosing the random excitation sequence long enough.} This result guarantees stability of the parameter estimation using this algorithm with an step size $\mu=n/m$.

Since the Bloch manifold is parametrized by only three parameters in $\Theta$ and thus has a low intrinsic dimension, we use the cover tree ANN searches to accelerate the projection step. We perform the following procedure iteratively until convergence:
\begin{align*}\label{eq:BLIP}
(X^l)^{k+1} = 
\Pp^{\epsilon}_{\text{cone}(\widetilde \Cc)} \left( (X^l)^{k} - \mu \Ff^H_{\Omega_l}\left(\Ff_{\Omega_i} ((X^l)^{k})  -Y^l\right) \right) 
\end{align*}
The approximate cone projection follows the update in~\eqref{eq:ANN1}.

Provided with the same embedding result (for the random EPI acquisition protocol) and according to Theorem~\ref{th:inexactLS2} we can deduce the linear convergence of this algorithm to the true solution for mildly-chosen approximations $\epsilon$ (and the same step size $\mu=n/m$) where the projection step has now the complexity $O(J\log(d))$. For large-sized dictionaries the gap between the computation costs of the exact and  inexact IPG algorithm can be significantly large. 

\section{Numerical Experiments}
\label{sec:expe}

\begin{figure}
	\centering
	\subfloat[Flip angles (excitations) sequence]{
		\includegraphics[width=.4\textwidth]{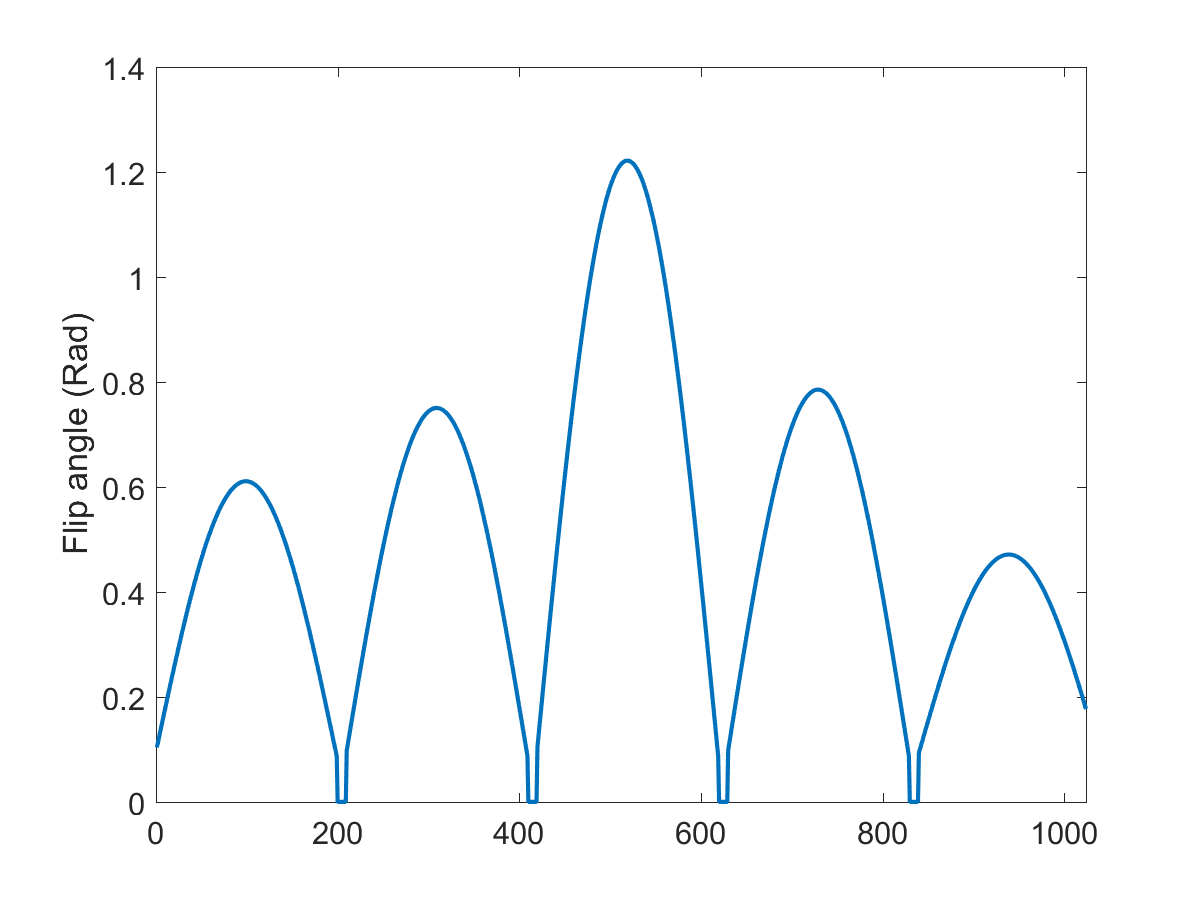}
	}
	\quad
	\subfloat[Normalized MRF dictionary atoms (the real components)]{
		\includegraphics[width=.4\textwidth]{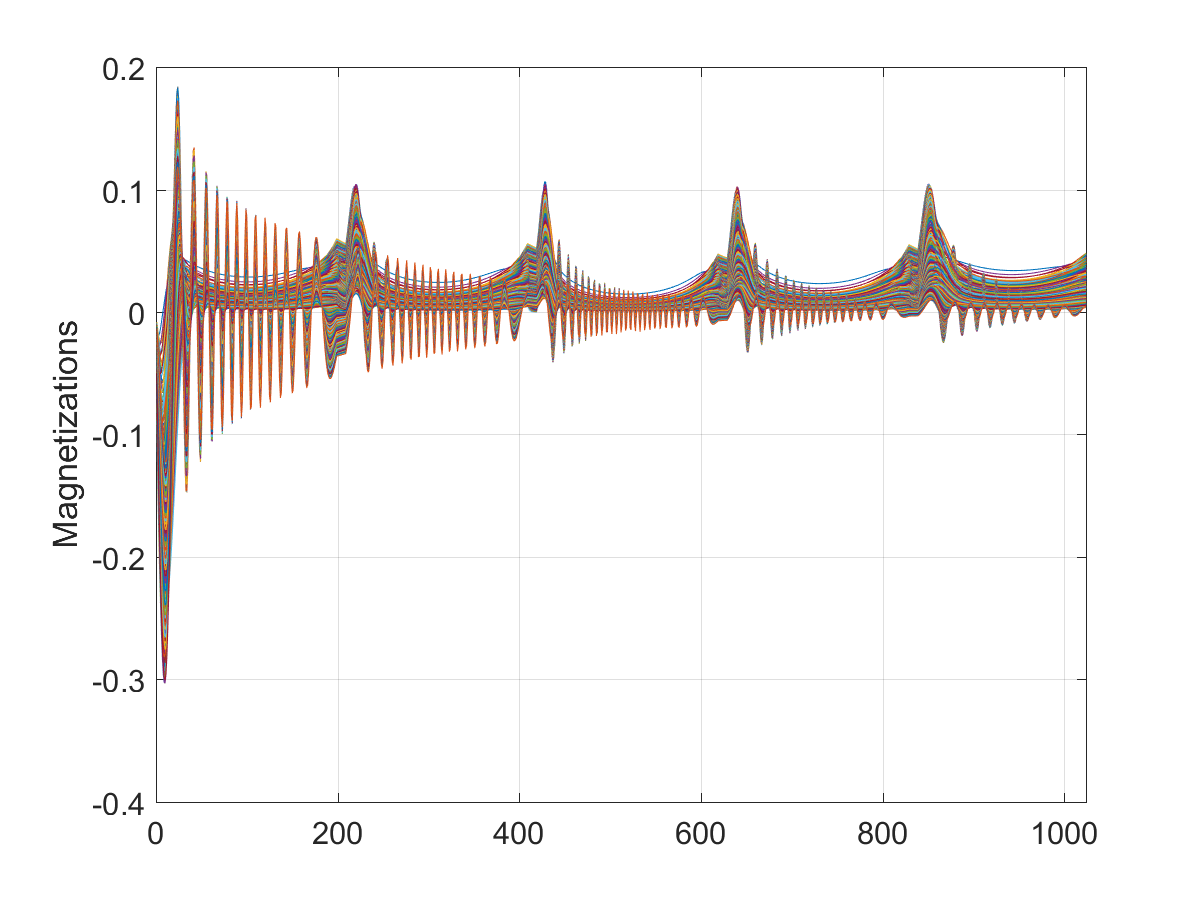}
	}
	\caption{TrueFISP flip angles and the corresponding normalized Bloch responses (fingerprints).\label{fig:FISP}}
\end{figure}

\begin{figure*}[!h]
	\centering
	\begin{minipage}{\textwidth}
		\centering
		\subfloat[Cover tree segments at scale 2]{\includegraphics[width=.22\textwidth]{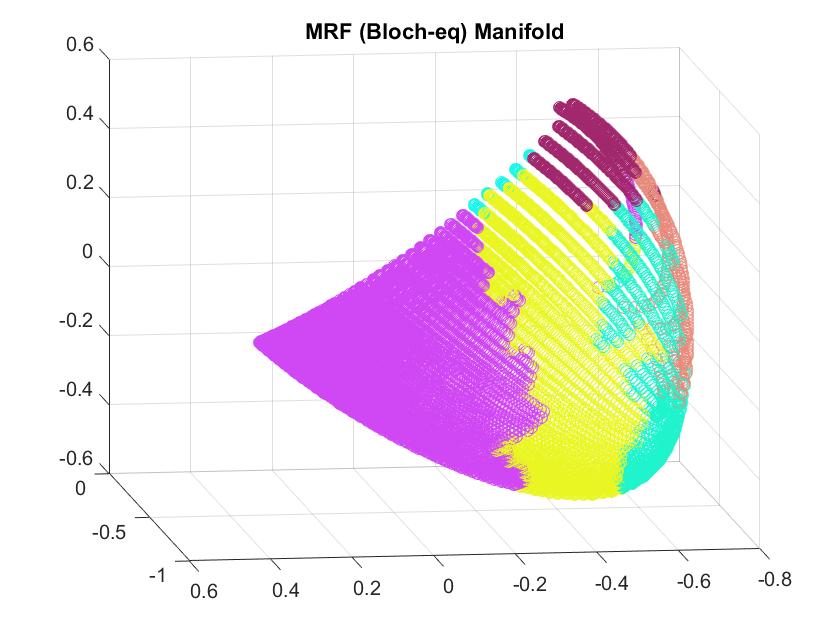} }
		\quad
		\subfloat[Cover tree segments at scale 3]{\includegraphics[width=.22\textwidth]{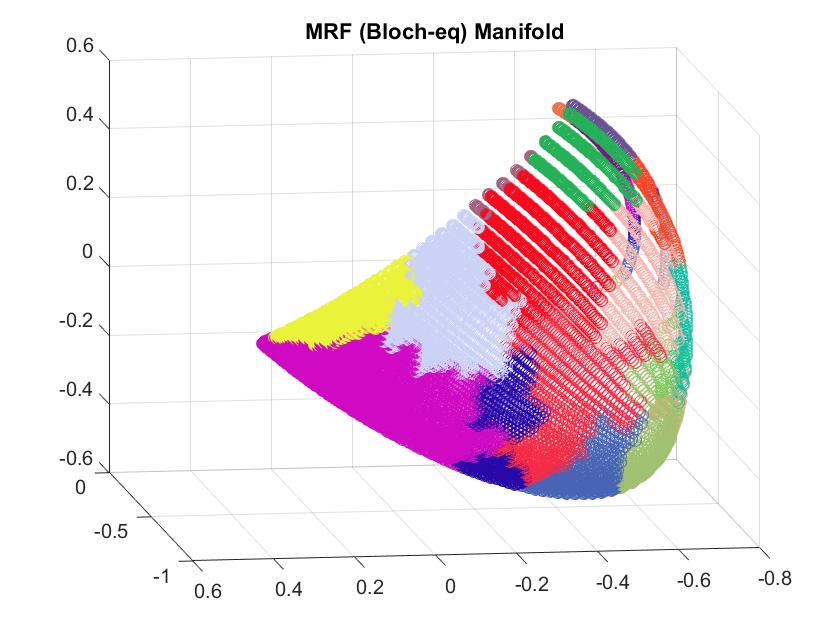} }
		\quad
		\subfloat[Cover tree segments at scale 4]{\includegraphics[width=.22\textwidth]{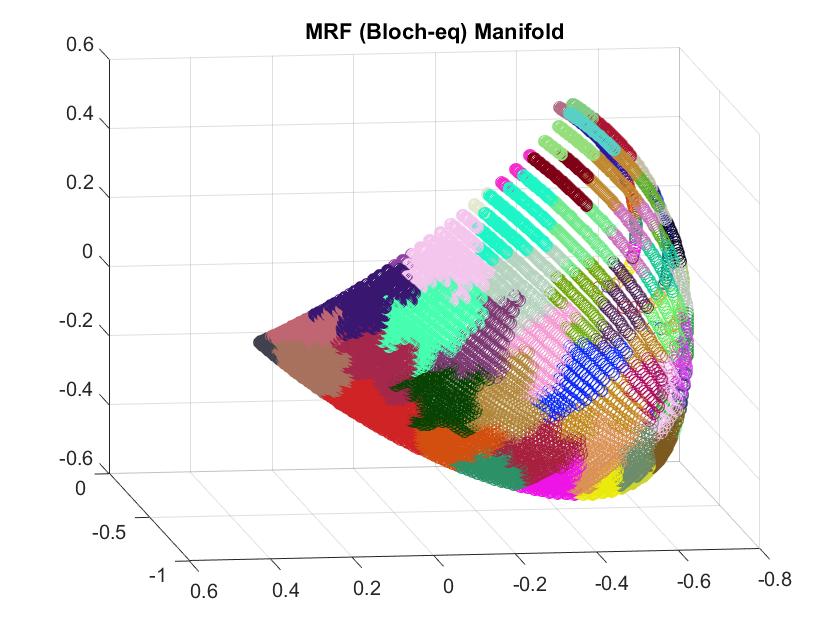} }
		\quad
		\subfloat[Cover tree segments at scale 5]{\includegraphics[width=.22\textwidth]{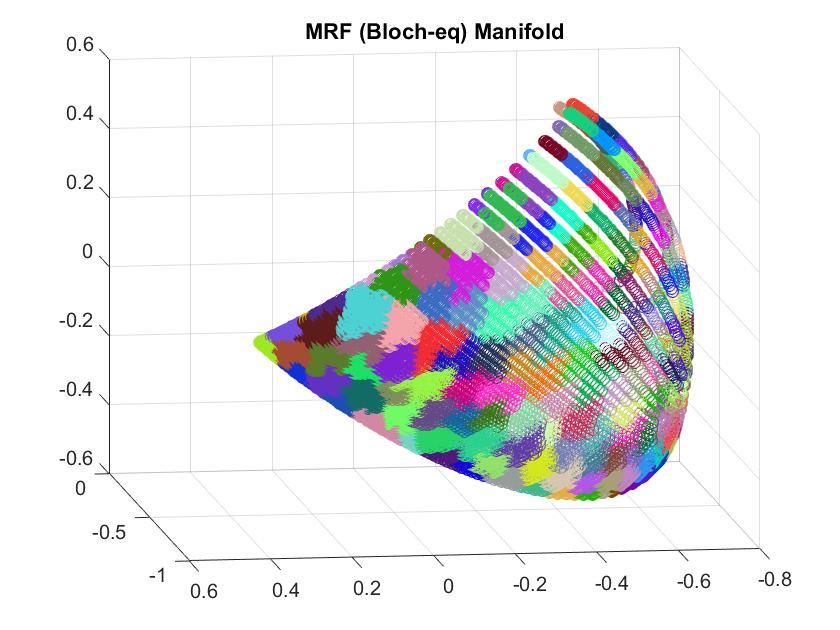} }
		
		\caption{A cover tree is formed on the Bloch response manifold composed of 14 scales: (a-d) data partitions are highlighted in different colours and demonstrated for scales 2-5. Low-scale partitions  divide into finer segments by traversing down the cover tree i.e. increasing the scale.  \label{fig:CT1} }
	\end{minipage}
\end{figure*}
We investigate application of our proposed scheme for accelerating the MRF reconstruction.
Parameter estimation consists of two off-line steps: i) forming the fingerprints dictionary  by solving the Bloch dynamic  equations for a wide range of parameters. 
ii) building a cover tree structure on the resulting dictionary which enables using fast ANN searches within the iterations of the inexact IPG algorithm. 

\begin{figure}
	\begin{minipage}{\linewidth}		
		\centering
		\subfloat[Brain phantom segments]{
		\includegraphics[width=.51\textwidth]{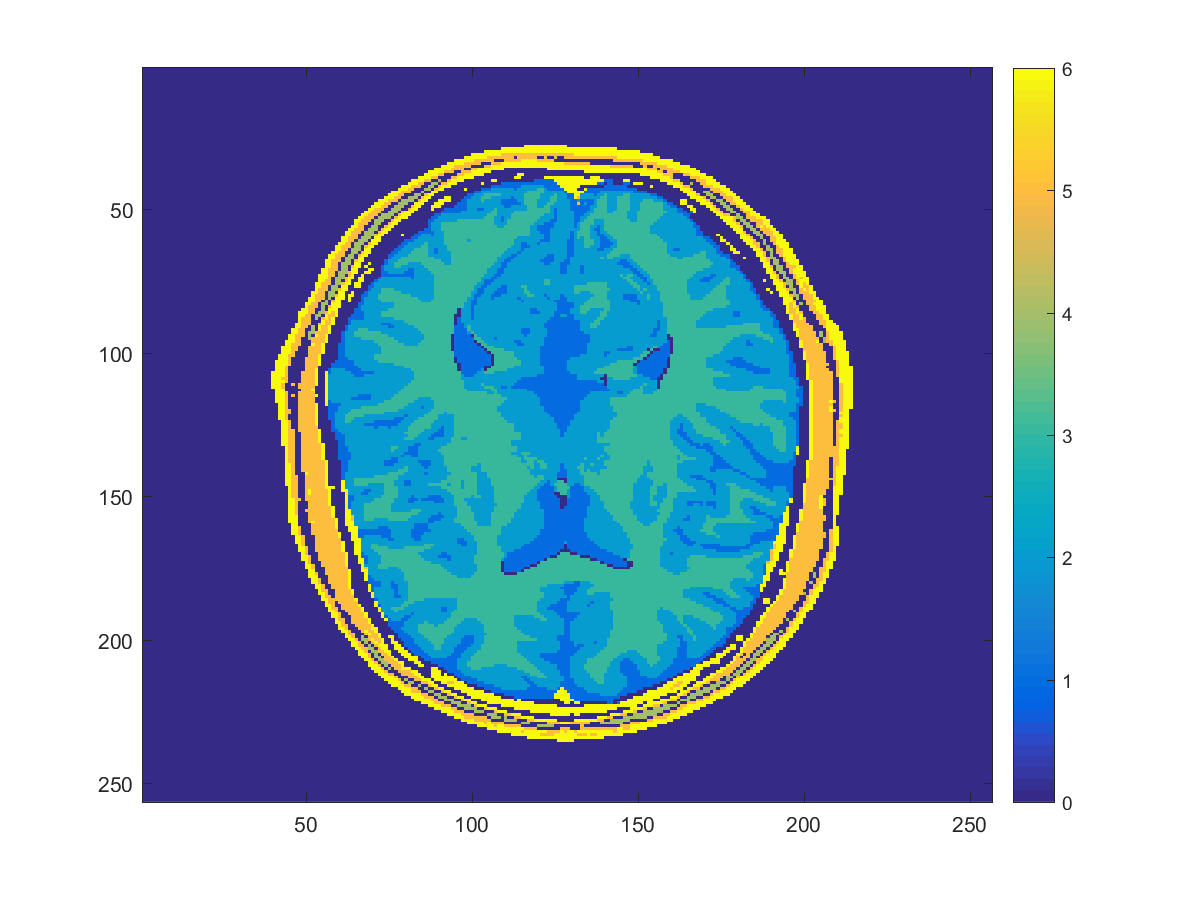}
	}
	\quad\hspace{-1cm}
	\subfloat[T1 map]{
		\includegraphics[width=.51\textwidth]{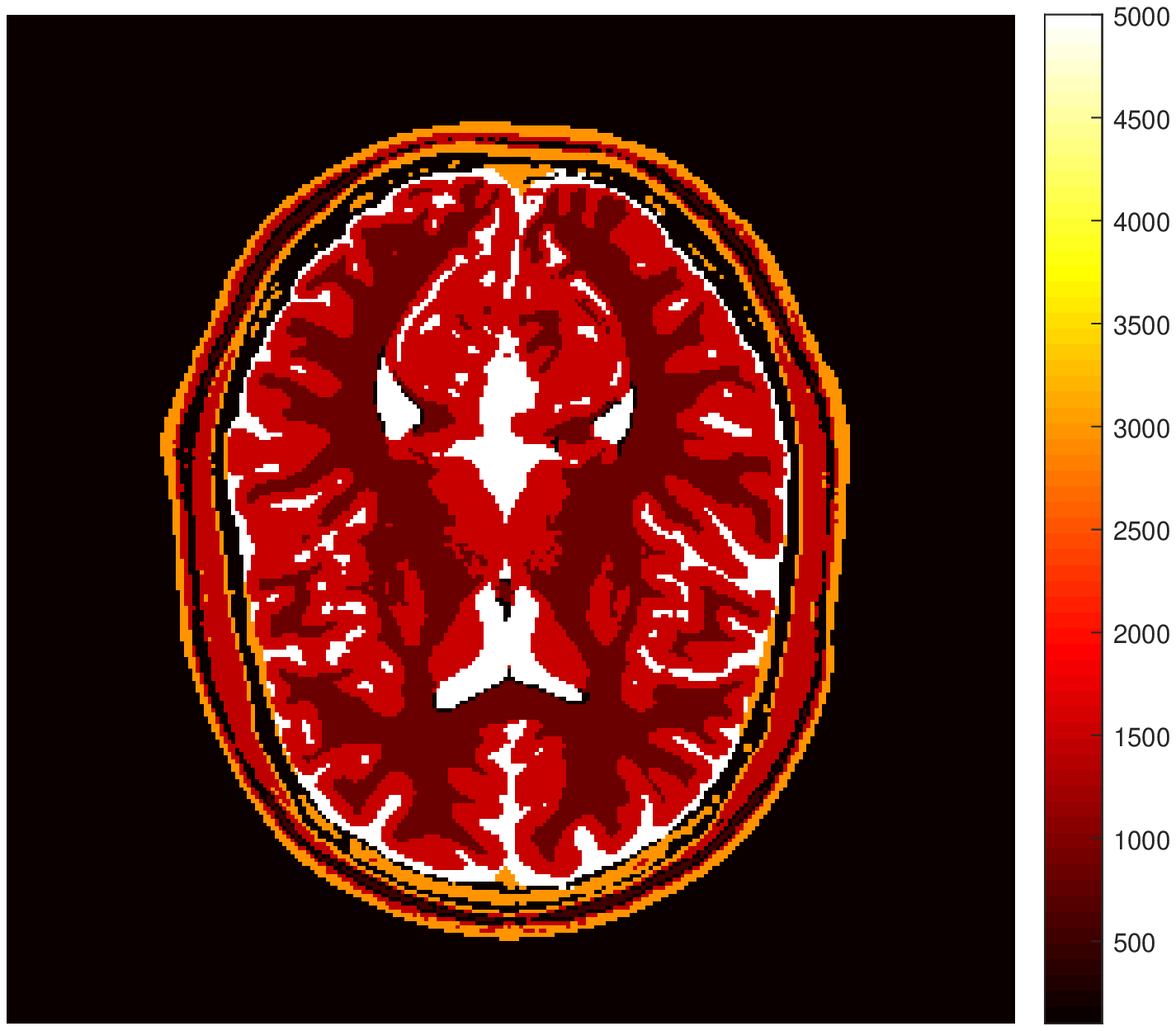}
	}
		\quad
		\subfloat[T2 map]{
			\includegraphics[width=.51\textwidth]{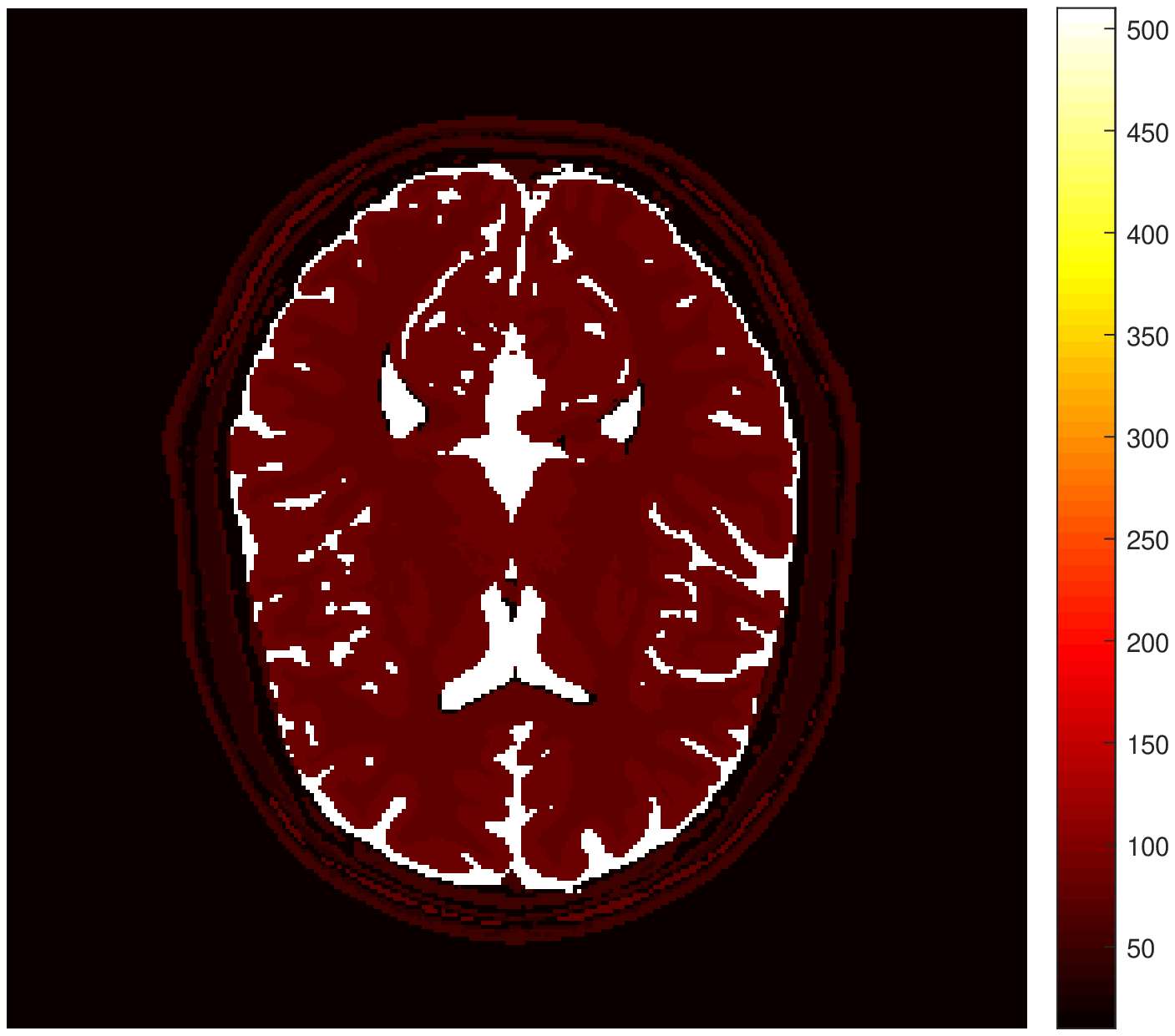}
		}
			\quad \hspace{-1cm}
			\subfloat[Proton density map]{
				\includegraphics[width=.51\textwidth]{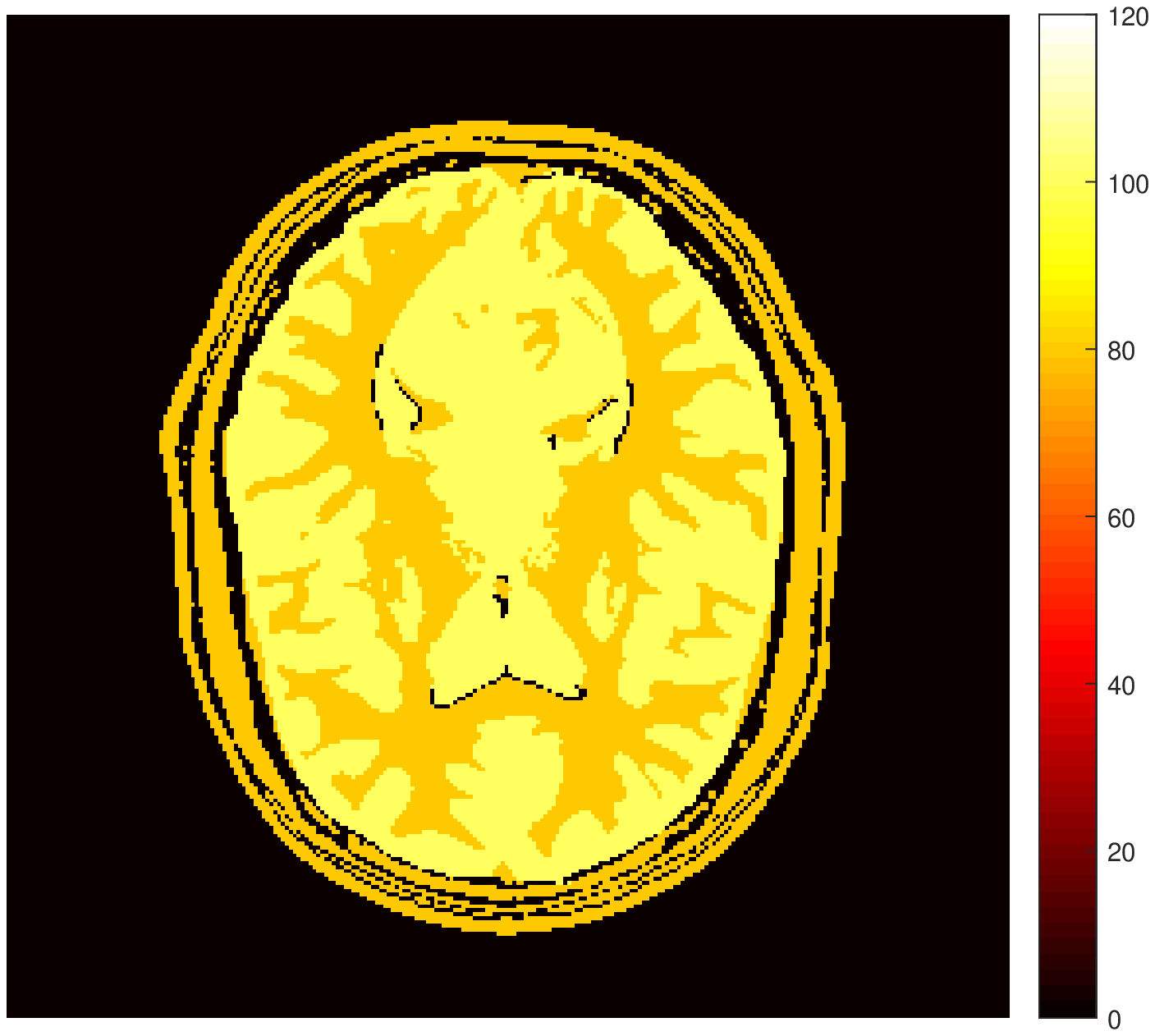}
			}
		\caption{Anatomical brain phantom. Segments correspond to Background, CSF, Grey Matter, White Matter, Muscle, Skin.\label{fig:phantom}}
	\end{minipage}
\end{figure}

We construct a cover tree on a dictionary composed of $d=48682$  fingerprints which are
sampled from the Bloch response of $d$ pairs of $(T1, T2)$ relaxation times ranging between 100-5000 (ms) and 20-1800 (ms), respectively.\footnote{In our experiments we set the off-resonance frequency parameter to zero, the  repetition time $TR=37$ (ms) and the echo time $TE=TR/2$.}	
Fingerprints $\psi_i\in \CC^{1024}$ are normalized magnetization responses of the corresponding parameters to a balanced
steady-state free-precession (bSSFP
or TrueFISP) sequence of  1024 slowly varying flip angles (excitations) ranging from $0^\circ$
to $60^\circ$, shown in Figure~\ref{fig:FISP}. From Figure~\ref{fig:CT1} we can observe the MRF low dimensional manifold across its first three principal components. The hierarchical data partitions resulted by the cover tree is also depicted for four scales (the tree has total 14 scales with the finest coverage resolution $\sim 10^{-4}$). The ground truth image is a synthetic $256\times256$ brain phantom with six $(T1,T2)$ segments previously used for evaluations in \cite{BLIPsiam}.  Figure~\ref{fig:phantom} shows the corresponding segments, $T1$, $T2$ and proton density maps. This phantom is synthesized with the corresponding Bloch responses and proton density according to model~\eqref{eq:immodel} to build the magnetization image $X$. 
The k-space subsampling is based on a lattice-based non random EPI protocol (i.e. uniform row subselection with a  shifted pattern for different image slices). We assume no noise on the measurements and we set the step size $\mu=n/m$ in all experiments.

\begin{figure*}[t]
	\centering
	\subfloat{\includegraphics[width=.3\linewidth]{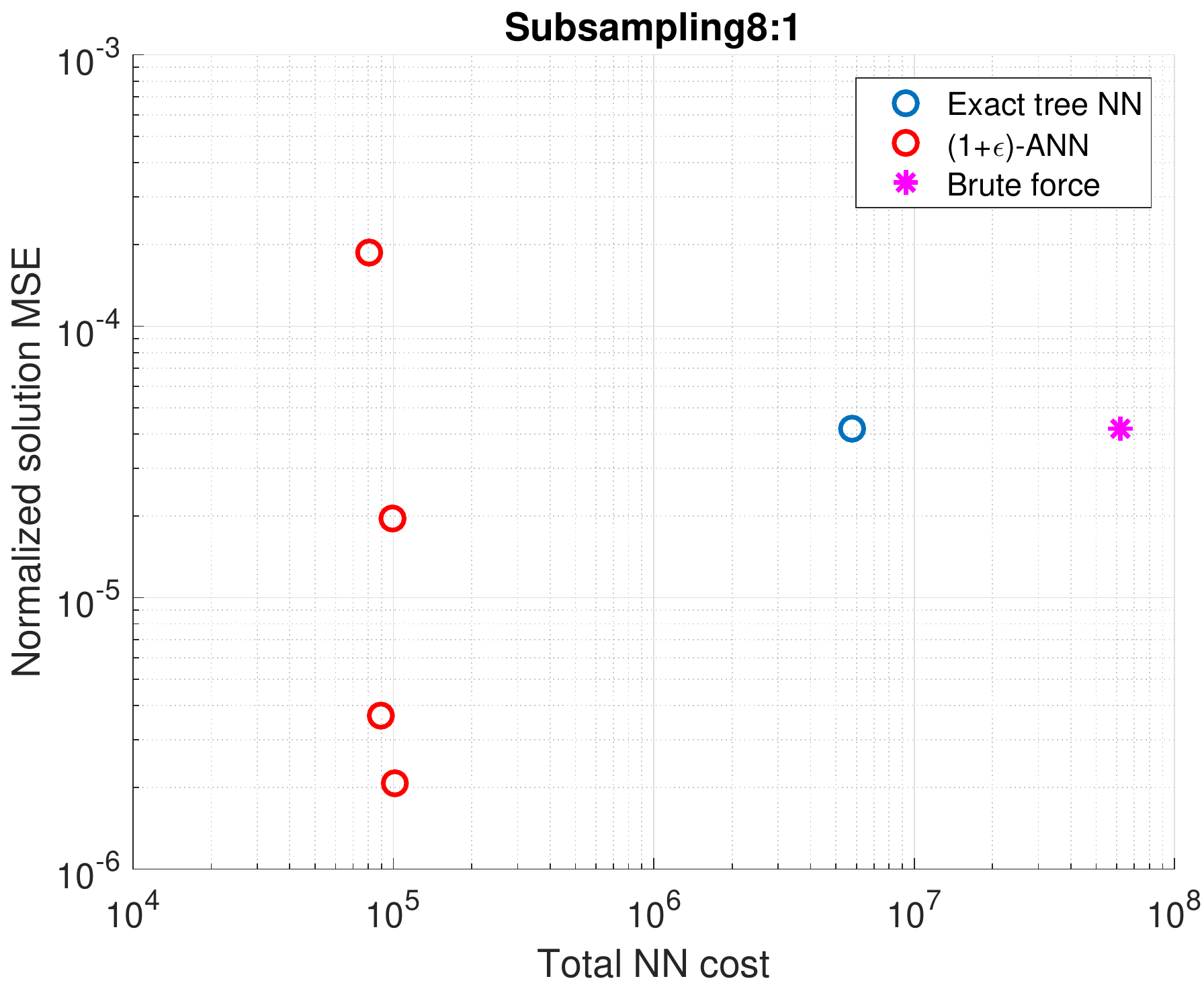} }
	\quad
	\subfloat{\includegraphics[width=.3\linewidth]{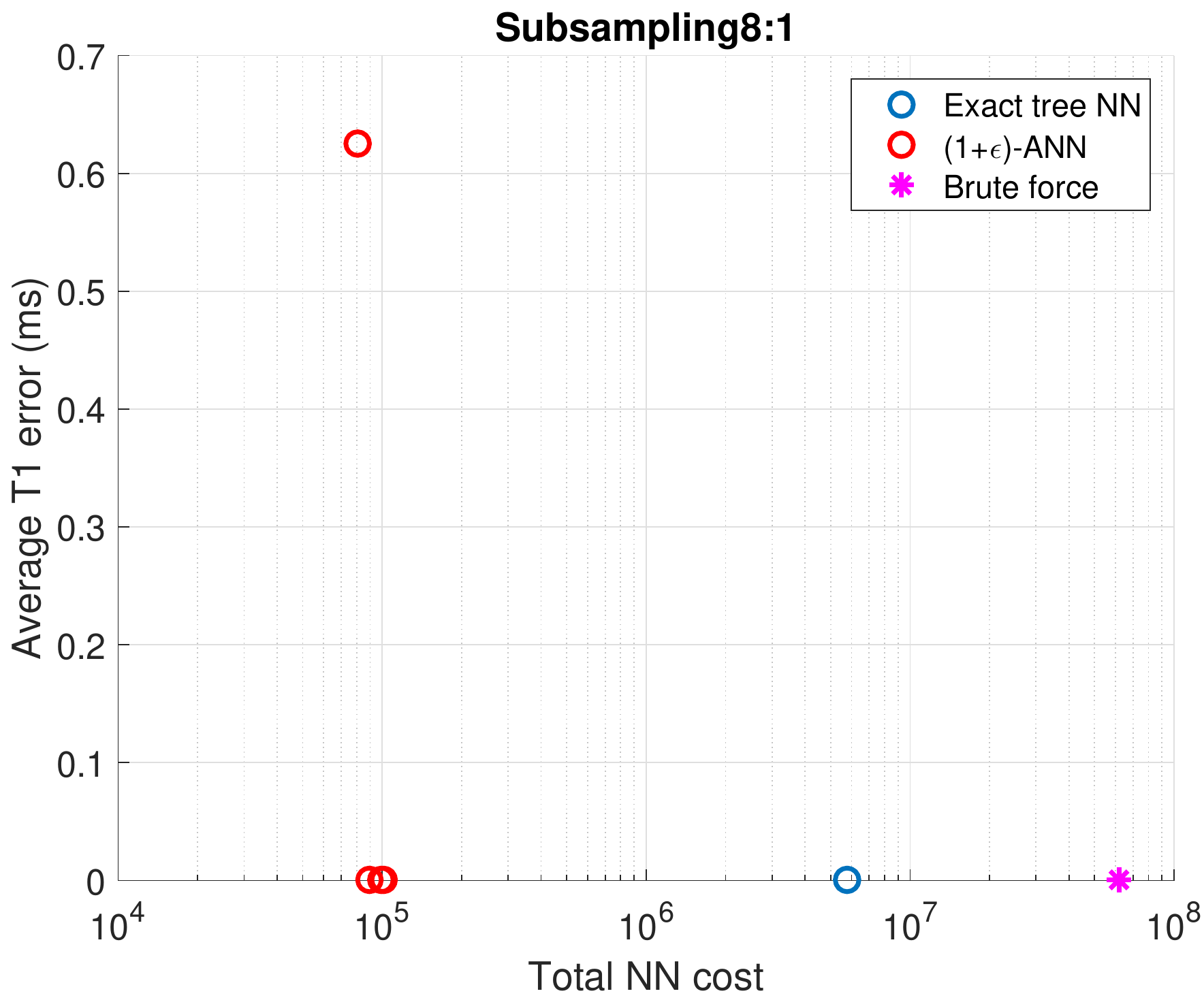} }
	\quad
	\subfloat{\includegraphics[width=.3\linewidth]{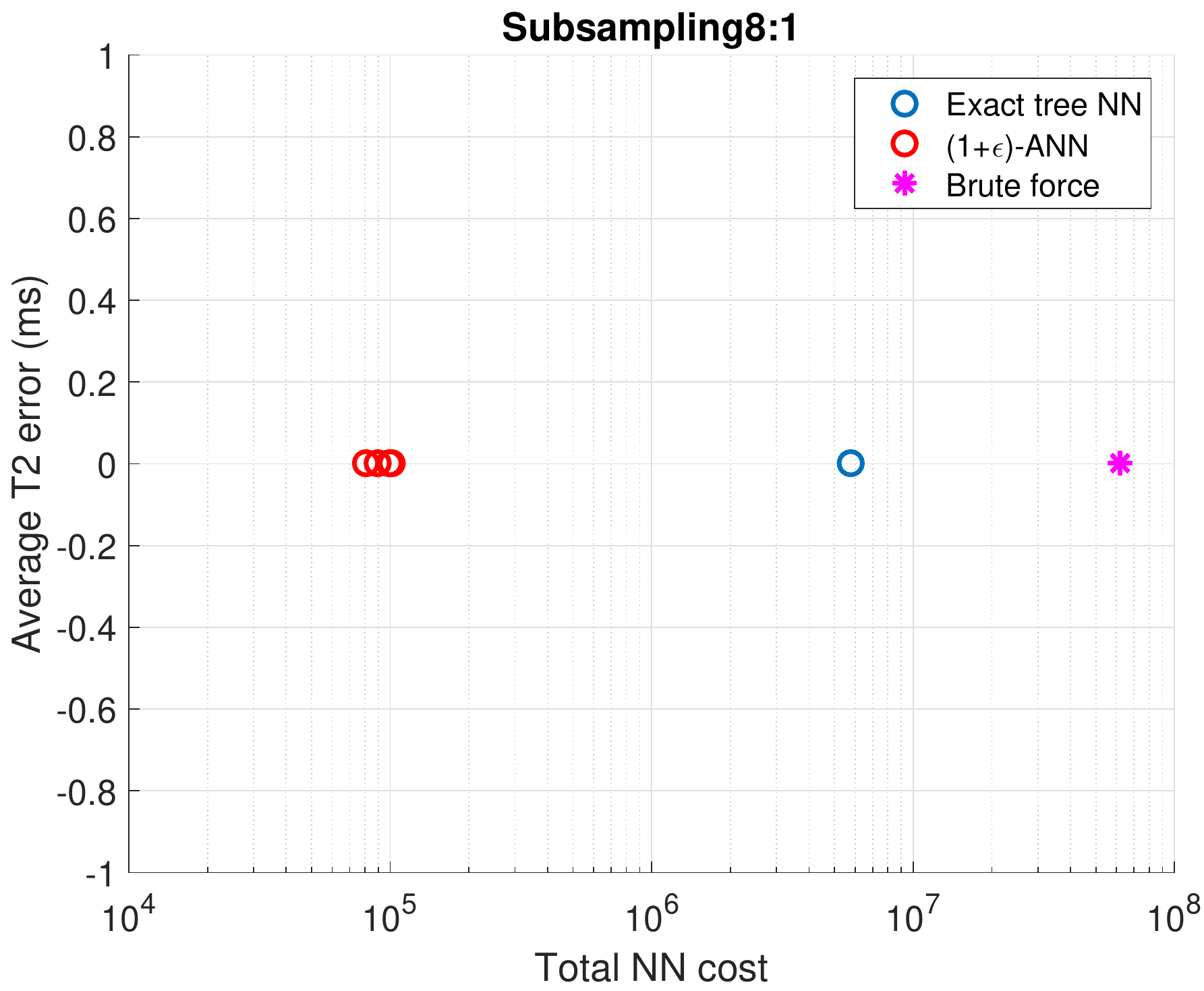} }
	
		\centering
		\subfloat{\includegraphics[width=.3\linewidth]{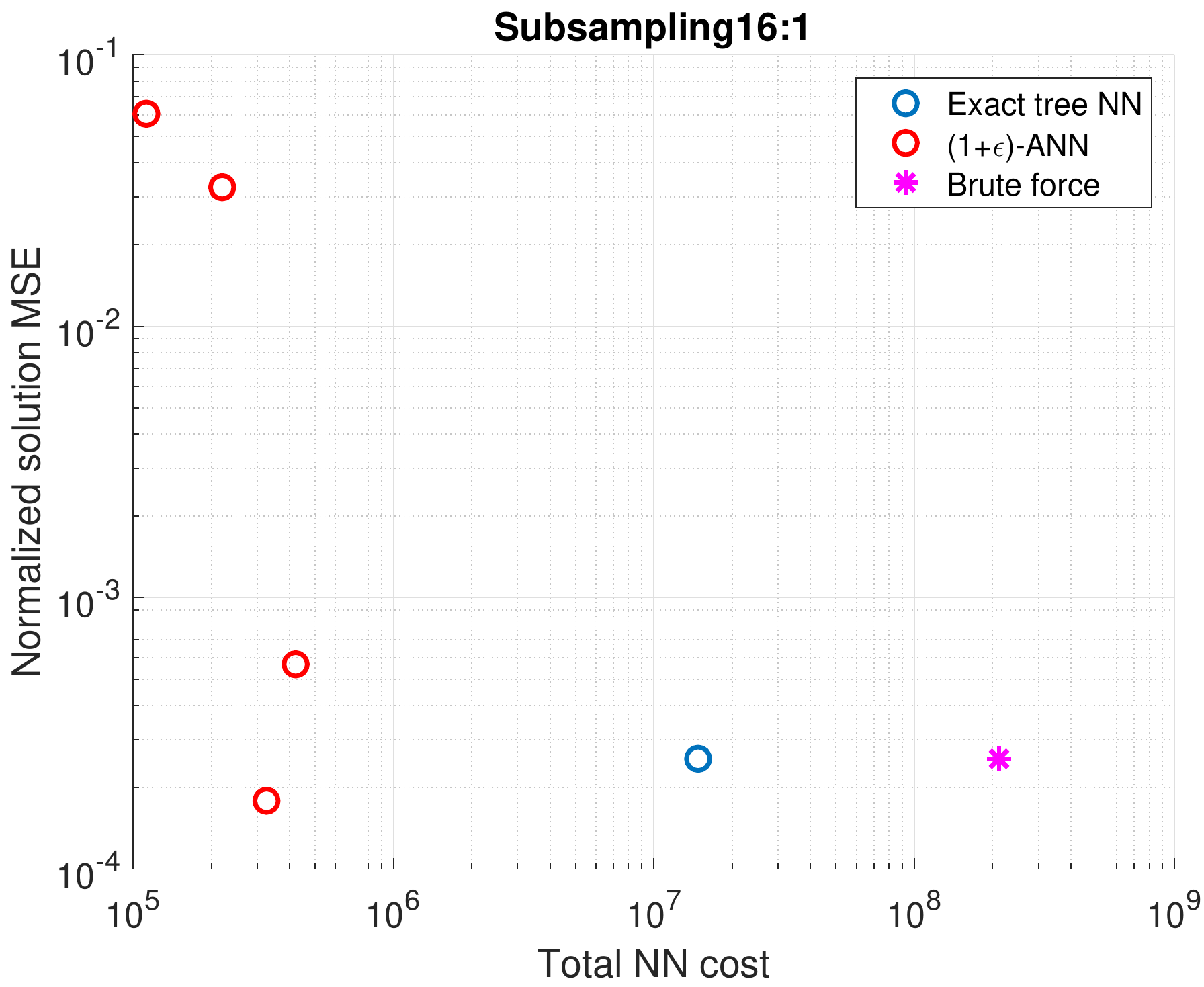} }
		\quad
		\subfloat{\includegraphics[width=.3\linewidth]{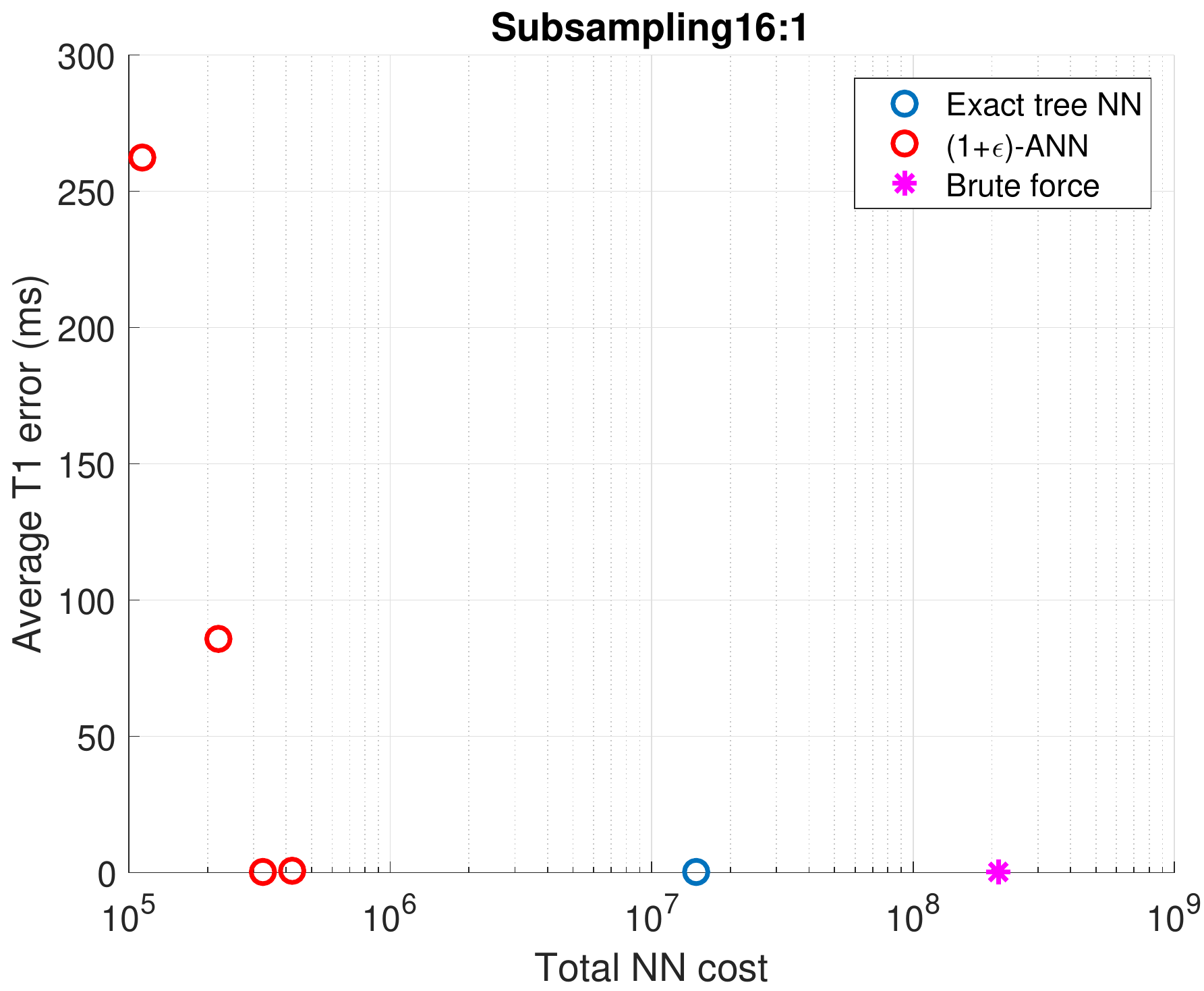} }
		\quad
		\subfloat{\includegraphics[width=.3\linewidth]{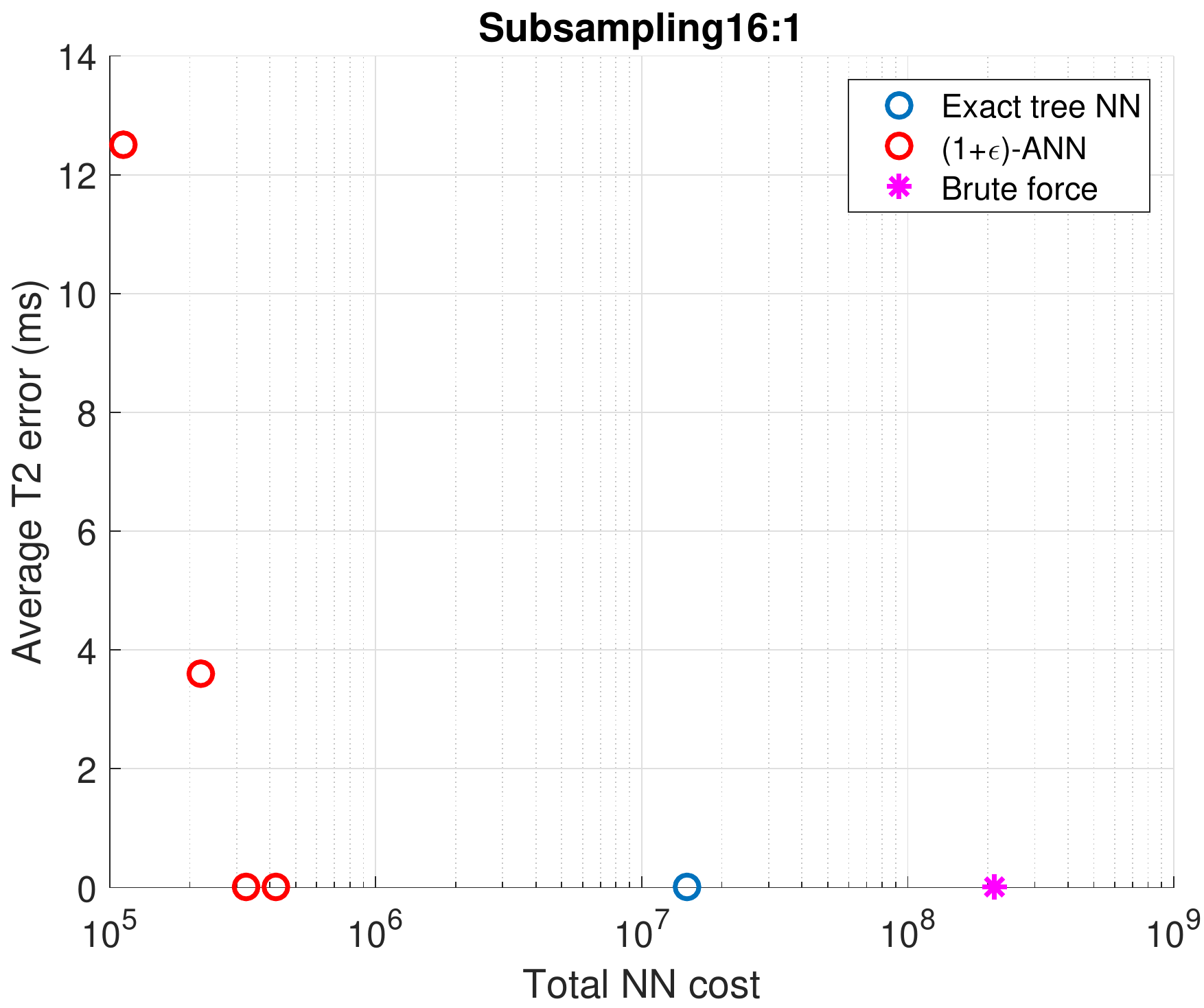} }
	
	\caption{Normalized solution MSE (i.e. $\frac{\norm{\hat x-x_0}}{\norm{x_0}}$) and  average T1, T2 errors (i.e. $\frac{1}{J}\sum_{j=1}^J |\hat T1(j) -T1(j)|$ and $\frac{1}{J}\sum_{j=1}^J |\hat T2(j) -T2(j)|$) vs. computations cost (projection step only), for two subsampling regimes. Here the  inexact IPG with $\epsilon=0.4$ achieves the lowest  solution MSE (for both subsampling regimes) among all tested methods with a clear computation cost advantage compared to the exact algorithms. 
	\label{fig:accu}}
\end{figure*}

We apply the $(1+\epsilon)$-ANN tree search for different values of $\epsilon=\{0, 0.2, 0.4, 0.6, 0.8\}$. 
We compare the performances of the exact and inexact data driven IPG on this synthetic phantom. 
We only report the cost of projections i.e. the total number of pairwise distances calculated for performing the NN or ANN searches.\footnote{Since the MRF's  forward/adjoint sampling operator is based on the Fast Fourier Transform, the gradient updates cost a tiny fraction of the search step  (particularly when dealing with large-size datasets) and thus we exclude them from our evaluations.}
Results are depicted in Figure~\ref{fig:accu} and indicate that the inexact algorithm (for a suitable parameter $\epsilon$) achieves a similar level of accuracy as for the exact IPG (with an exhaustive search) by saving 2-3 orders of magnitude in computations. Using an exact tree search (i.e. $\epsilon=0$) also saves computations by an order of magnitude compared to using a brute-force search.  
Remarkably, for subsampling ratio $8:1$ the solution accuracy of the approximate scheme is about an order of magnitude better than the exact IPG. Although our theoretical results do not cover such observation, we shall relate it to a common practical knowledge that using soft decisions (e.g. here approximations)  generally improves the performance of nonconvex algorithms compared to making hard decisions, and introduces a notion of robustness against undesirable local minima in such settings.

\section{Conclusions}
We proposed an inexact IPG algorithm for data driven CS problem with an application to fast MRF parameter estimation. The projection step is implemented approximately by  using cover trees ANN searches.  Thanks to the IPG's robustness against approximations, by using this algorithm one achieves a  comparable (and sometimes empirically enhanced) solution precision to the exact algorithm  however with significantly less amount of  computations regarding iteratively searching through large datasets.

\bibliographystyle{IEEEtran}
\bibliography{mybiblio}

\begin{thebibliography}{10}
\providecommand{\url}[1]{#1}
\csname url@samestyle\endcsname
\providecommand{\newblock}{\relax}
\providecommand{\bibinfo}[2]{#2}
\providecommand{\BIBentrySTDinterwordspacing}{\spaceskip=0pt\relax}
\providecommand{\BIBentryALTinterwordstretchfactor}{4}
\providecommand{\BIBentryALTinterwordspacing}{\spaceskip=\fontdimen2\font plus
\BIBentryALTinterwordstretchfactor\fontdimen3\font minus
  \fontdimen4\font\relax}
\providecommand{\BIBforeignlanguage}[2]{{%
\expandafter\ifx\csname l@#1\endcsname\relax
\typeout{** WARNING: IEEEtran.bst: No hyphenation pattern has been}%
\typeout{** loaded for the language `#1'. Using the pattern for}%
\typeout{** the default language instead.}%
\else
\language=\csname l@#1\endcsname
\fi
#2}}
\providecommand{\BIBdecl}{\relax}
\BIBdecl

\bibitem{DonohoCS}
\BIBentryALTinterwordspacing
D.~L. Donoho, ``Compressed sensing,'' \emph{IEEE Transactions on Information
  Theory}, vol.~52, no.~4, pp. 1289--1306, Apr. 2006.
\BIBentrySTDinterwordspacing

\bibitem{CRT:CS}
E.~J. Candes, J.~Romberg, and T.~Tao, ``Robust uncertainty principles: exact
  signal reconstruction from highly incomplete frequency information,''
  \emph{IEEE Transactions on Information Theory}, vol.~52, no.~2, pp. 489--509,
  Feb 2006.

\bibitem{RichCSreview}
R.~G. Baraniuk, V.~Cevher, and M.~B. Wakin, ``Low-dimensional models for
  dimensionality reduction and signal recovery: A geometric perspective,''
  \emph{Proceedings of the IEEE}, vol.~98, no.~6, pp. 959--971, June 2010.

\bibitem{Volkan:bigdata}
V.~Cevher, S.~Becker, and M.~Schmidt, ``Convex optimization for big data:
  Scalable, randomized, and parallel algorithms for big data analytics,''
  \emph{IEEE Signal Processing Magazine}, vol.~31, no.~5, pp. 32--43, Sept
  2014.

\bibitem{Blumen}
T.~Blumensath, ``Sampling and reconstructing signals from a union of linear
  subspaces,'' \emph{IEEE Transactions on Information Theory}, vol.~57, no.~7,
  pp. 4660--4671, July 2011.

\bibitem{recht:discretize}
G.~Tang, B.~N. Bhaskar, and B.~Recht, ``Sparse recovery over continuous
  dictionaries-just discretize,'' in \emph{2013 Asilomar Conference on Signals,
  Systems and Computers}, 2013, pp. 1043--1047.

\bibitem{SCOOP}
\BIBentryALTinterwordspacing
M.~Golbabaee, A.~Alahi, and P.~Vandergheynst, ``Scoop: A real-time sparsity
  driven people localization algorithm,'' \emph{Journal of Mathematical Imaging
  and Vision}, vol.~48, no.~1, pp. 160--175, Jan. 2014.
\BIBentrySTDinterwordspacing

\bibitem{me:inexactIPG}
M.~Golbabaee and M.~E. Davies, ``Inexact gradient projection and fast data
  driven compressed sensing,'' \emph{arXiv preprint arXiv:1706.00092}, 2017.

\bibitem{beygelzimer2006cover}
A.~Beygelzimer, S.~Kakade, and J.~Langford, ``Cover trees for nearest
  neighbor,'' in \emph{Proceedings of the 23rd international conference on
  Machine learning}.\hskip 1em plus 0.5em minus 0.4em\relax ACM, 2006, pp.
  97--104.

\bibitem{Navigating}
R.~Krauthgamer and J.~R. Lee, ``Navigating nets: Simple algorithms for
  proximity search,'' in \emph{Proceedings of the Fifteenth Annual ACM-SIAM
  Symposium on Discrete Algorithms}, ser. SODA '04, 2004.

\bibitem{MRF}
D.~Ma, V.~Gulani, N.~Seiberlich, K.~Liu, J.~Sunshine, J.~Durek, and
  M.~Griswold, ``Magnetic resonance fingerprinting,'' \emph{Nature}, vol. 495,
  no. 7440, pp. 187--192, 2013.

\bibitem{BLIPsiam}
M.~Davies, G.~Puy, P.~Vandergheynst, and Y.~Wiaux, ``A compressed sensing
  framework for magnetic resonance fingerprinting,'' \emph{SIAM Journal on
  Imaging Sciences}, vol.~7, no.~4, pp. 2623--2656, 2014.

\bibitem{BW:manifold}
\BIBentryALTinterwordspacing
R.~G. Baraniuk and M.~B. Wakin, ``Random projections of smooth manifolds,''
  \emph{Foundations of Computational Mathematics}, vol.~9, no.~1, pp. 51--77,
  2009.
\BIBentrySTDinterwordspacing

\bibitem{TIPHSI}
M.~Golbabaee, S.~Arberet, and P.~Vandergheynst, ``Compressive source
  separation: Theory and methods for hyperspectral imaging,'' \emph{IEEE
  Transactions on Image Processing}, vol.~22, no.~12, pp. 5096--5110, Dec 2013.

\bibitem{Kobarg2014}
\BIBentryALTinterwordspacing
J.~H. Kobarg, P.~Maass, J.~Oetjen, O.~Tropp, E.~Hirsch, C.~Sagiv, M.~Golbabaee,
  and P.~Vandergheynst, ``Numerical experiments with maldi imaging data,''
  \emph{Advances in Computational Mathematics}, vol.~40, no.~3, pp. 667--682,
  2014.
\BIBentrySTDinterwordspacing

\bibitem{meEUSIPCO10}
M.~Golbabaee, S.~Arberet, and P.~Vandergheynst, ``Multichannel compressed
  sensing via source separation for hyperspectral images,'' in \emph{2010 18th
  European Signal Processing Conference}, 2010, pp. 1326--1329.

\bibitem{dasgupta2008}
S.~Dasgupta and Y.~Freund, ``Random projection trees and low dimensional
  manifolds,'' in \emph{Proceedings of the fortieth annual ACM symposium on
  Theory of computing}.\hskip 1em plus 0.5em minus 0.4em\relax ACM, 2008, pp.
  537--546.

\bibitem{toftsqmri}
P.~Tofts, \emph{Quantitative MRI of the brain: measuring changes caused by
  disease}.\hskip 1em plus 0.5em minus 0.4em\relax John Wiley \& Sons, 2005.

\end{thebibliography}

\end{document}